\documentclass[11pt]{article}

\usepackage[final]{acl}

\usepackage{times}
\usepackage{latexsym}

\usepackage[T1]{fontenc}

\usepackage[utf8]{inputenc}

\usepackage{microtype}

\usepackage{inconsolata}

\usepackage{graphicx}
\usepackage{booktabs}
\usepackage[section]{placeins}

\usepackage{amsmath}

\makeatletter
\@ifundefined{linenomath*}{\newenvironment{linenomath*}{}{}}{}
\makeatother

\title{On the Structure of Address in Multi-Party Dialogue:\\ From Discrete Labels to Continuous Levels}

\author{
 \textbf{Taiga Mori},
 \textbf{Koji Inoue},
 \textbf{Divesh Lala},
 \textbf{Tatsuya Kawahara}
\\
\\
 Graduate School of Informatics, Kyoto University, Japan
\\
 \small{
   \textbf{Correspondence:} \href{mori@sap.ist.i.kyoto-u.ac.jp}{mori@sap.ist.i.kyoto-u.ac.jp}
 }
}

\begin{document}
\maketitle

\begin{abstract}
In multi-party dialogues between a dialogue system and multiple users, 
identifying to whom an utterance is addressed is a key challenge. 
Prior work has typically treated addressee detection as a multi-class classification task, 
selecting a single label representing an individual participant or the group. 
This formulation assumes that address is inherently discrete 
and has primarily been used for predicting turn-taking. 
In this paper, we revisit this assumption by analyzing address as a continuous phenomenon. 
Using a multi-party human dialogue corpus annotated by multiple annotators, 
we construct both binary address labels derived from majority-vote addressee labels 
and continuous address levels inferred from annotator judgments using a latent-variable model. 
We then examine how these representations relate to turn-taking as well as listener behaviors, 
including gaze and backchannels. 
Our results show that, in addition to turn-taking, both gaze and backchannels are associated with address. 
Furthermore, models using continuous address levels achieve better predictive fit than those using discrete labels, 
suggesting that address may exhibit graded structure. 
Finally, we discuss the future directions of addressee detection research 
based on the findings of this study.
\end{abstract}

\section{Introduction}

Identifying to whom an utterance is addressed 
is one of the fundamental challenges 
in multi-party dialogues involving multiple users 
and a dialogue system. As discussed below, 
most prior work on addressee detection 
has formulated addressee inference as a classification problem. 
In human-human-computer interaction, 
the task is typically treated as a binary classification problem that 
determines whether a user's utterance is 
addressed to the system or to another human participant. 
In multi-party human dialogue, 
by contrast, 
it is commonly formulated as selecting a single addressee 
from among labels corresponding to individual participants or the group as a whole. 
Moreover, 
in dialogue-system research, 
addressee detection has primarily been used for predicting turn-taking.

Such conventional formulations rely on the simplified assumption that 
each utterance is directed toward a single discrete addressee or 
toward a predefined category such as the whole group. 
However, there is little reason to assume that naturally occurring dialogue is always so simple, 
and address may in fact be more complex. 
Under such formulations, 
disagreement among annotators is typically treated as mere error. 
In addition, 
address is often modeled as an isolated phenomenon, 
and relatively little work has examined its relationship to listener behaviors that 
are likely to be associated with it, 
or attempted to model these phenomena jointly.

From this perspective, 
the present study analyzes addressee annotations assigned 
by multiple annotators to multi-party human dialogues 
and shows that address may not be a purely discrete phenomenon, 
but instead may exhibit graded structure. 
We also introduce \textit{address level} as a new concept 
for addressee modeling, 
representing addressivity as a continuous property. 
Furthermore, we show that address is related not only to turn-taking 
but also to other listener behaviors.

The contributions of this study are threefold. 
First, we provide empirical grounds 
for reformulating addressee modeling not as a discrete classification problem 
but as a continuous estimation problem. 
Second, we propose an operationalization that 
infers participant-wise continuous representations 
from categorical annotations. 
Third, we show that these continuous addressivity measures 
correlate with observable listener behaviors, 
thereby supporting their validity 
as a representation of interactional structure.

This paper is structured as follows. 
Section 2 organizes previous research on addressee detection 
and situates the present study within that literature. 
Section 3 describes the dataset and annotation scheme 
and presents a preliminary analysis of the ambiguity inherent in addressee annotation. 
Section 4 explains how address labels and address levels were derived, 
as well as how turn-taking, listener gaze, and backchannel behavior were modeled. 
Section 5 presents the quantitative results and a qualitative analysis. 
Section 6 concludes the paper with a discussion of implications and future directions. 
Limitations are discussed in a separate section.

\section{Related Work}

Most prior work on addressee detection has formulated the task 
as a discrete classification problem. 
Early studies on meeting dialogues treated addressee detection 
as assigning discrete labels to dialogue acts, 
distinguishing utterances addressed to an individual participant 
from those addressed to the group; 
when an individual was addressed, 
the label specified the participant ID 
\citep{jovanovic2006addressee,op-den-akker-op-den-akker-2009-addressed}. 
In text-based multi-party dialogue research, 
the task has often been formulated as selecting one addressee 
from the candidate interlocutors appearing in the dialogue context 
\citep{ouchi2016addressee,zhang2018addressee,sato2018addressee,gu2021mpc,zhu2023robust}. 
In human-human-computer settings, 
by contrast, 
the label space is often binary, 
distinguishing system-addressed speech from speech addressed to another human participant 
\citep{baba2011identifying,shriberg2012learning,tsai2015study,nakano2013implementation}.

Regarding the timing of prediction, 
many studies perform inference after the target turn has been fully observed. 
Early work treated the task as assigning addressee labels to dialogue acts or manually segmented utterances 
\citep{jovanovic2006addressee}, 
and this utterance-level formulation remains common in recent studies. 
For example, 
\citet{ouchi2016addressee} jointly predict the addressee 
and response given the full context and target utterance, 
while \citet{penzo2024llms} evaluate addressee recognition for the final turn in multi-party dialogues. 
Thus, a common paradigm is post-hoc prediction at the utterance level, 
rather than incremental estimation during the progression of an utterance.

Furthermore, 
addressee detection has often been treated as an auxiliary task 
for turn-taking or response selection. 
In human-human-computer interaction, 
it is used to determine whether the system should respond to an utterance 
\citep{shriberg2012learning,tsai2015study}. 
In text-based multi-party dialogue, 
addressee selection is commonly integrated with response selection 
\citep{ouchi2016addressee,zhang2018addressee}, 
and MPC-BERT \citep{gu2021mpc} also treats addressee recognition 
as one component among multiple dialogue understanding tasks. 
In this way, 
addressee detection has often been used as an intermediate representation 
for turn-taking and response generation, 
rather than being studied as an independent dialogue behavior.

From a methodological perspective, 
the field has undergone a progressive evolution. 
Early studies primarily relied on statistical methods such as SVMs and logistic regression 
using features derived from gaze, prosody, 
and lexical cues 
\citep{jovanovic2006addressee,shriberg2012learning,tsai2015study}. 
Subsequently, 
neural network-based approaches were introduced to model dialogue context and speaker relationships 
\citep{ravuri2014neural,ouchi2016addressee,zhang2018addressee}. 
Later, multimodal deep learning approaches incorporating visual and contextual information were proposed 
\citep{akhtiamov2017speech,le2018deep}. 
More recently, 
recent work has also explored pretrained language models \citep{gu2021mpc} 
and inference-based approaches using LLMs and MLLMs 
\citep{zhu2023robust,penzo2024llms,inoue2025llm,mori2026analysing}.

Along with these methodological advances, 
predictive performance has steadily improved. 
While early studies reported accuracies of around 80\% \citep{jovanovic2006addressee}, 
recent approaches using MLLMs have achieved substantially higher performance. 
For instance, \citet{mori2026analysing} reports a Macro-F1 score exceeding 0.9 for cases 
where the single next speaker is identifiable. 
Moreover, as performance improves, 
recent research has begun to address additional challenges 
such as multilingual settings \citep{sato2018addressee} 
and robustness under noisy conditions \citep{zhu2023robust}.

In this study, 
we revisit these underlying assumptions 
by analyzing addressee annotations from multiple annotators, 
as well as the relationships between addressing behavior, 
turn-taking, listener gaze, and backchannels. 
Based on these analyses, 
we discuss future directions for addressee detection research.

\section{Dataset}
\subsection{Corpus Description}

\begin{figure}[tb]
  \centering
  \includegraphics[width=\columnwidth]{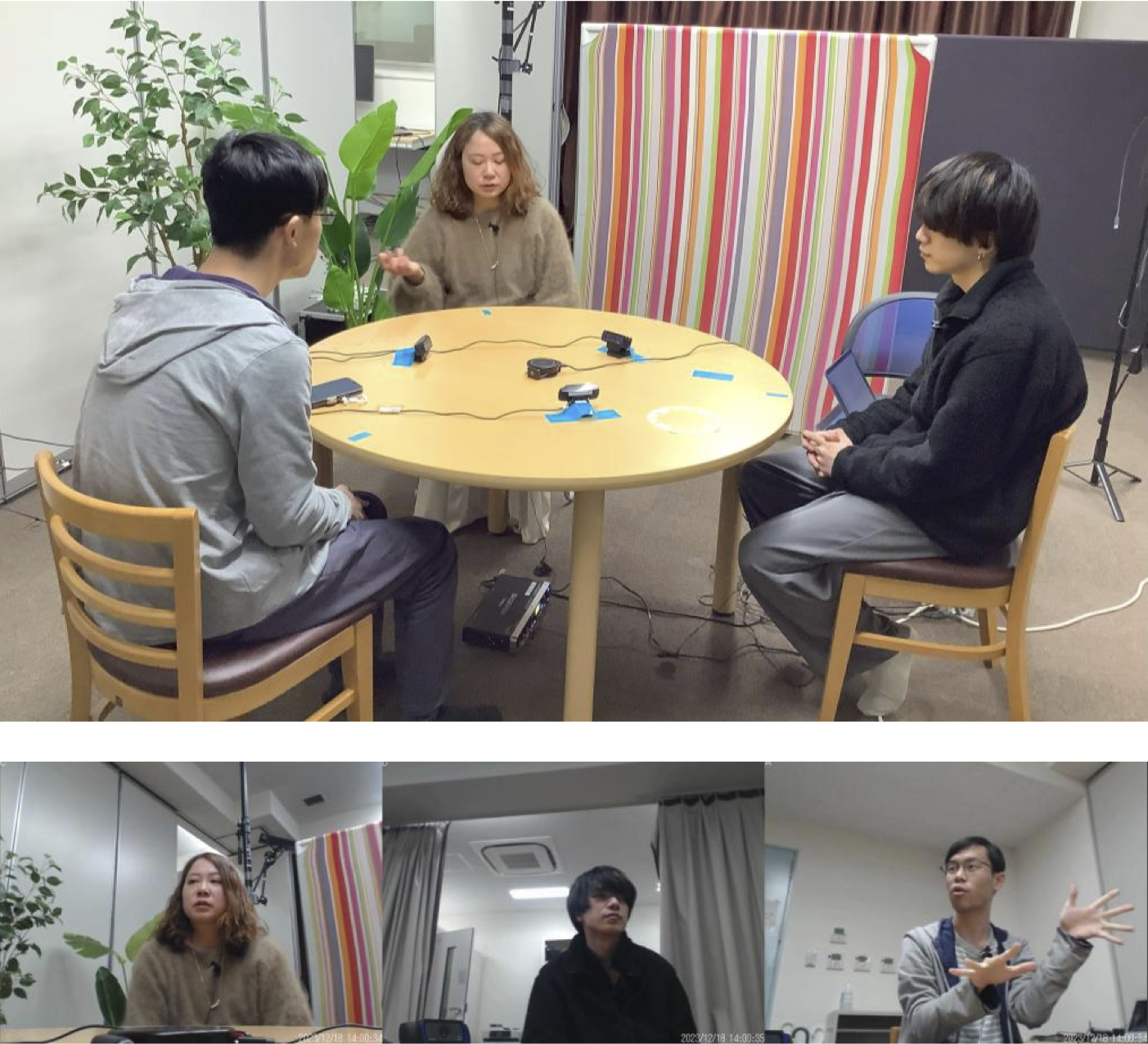}
  \caption{A snapshot from the Teidan corpus.}
  \label{fig:TEIDAN}
\end{figure}

In this study, we used the dataset introduced in prior work by \citet{inoue2025llm}. 
This corpus contains casual discussions 
in Japanese among triads of members of the same laboratory.
Twelve groups each conversed on three topics, 
resulting in a total of 36 recorded dialogues. 
Figure~\ref{fig:TEIDAN} shows a screenshot 
from one of the recorded dialogues.
The average duration of a dialogue is 6 minutes and 4 seconds, 
and the total duration is 3 hours, 38 minutes, and 27 seconds. 
The data are annotated manually with turn and backchannel information. 
A turn is defined as a segment during which one participant speaks continuously. 
By contrast, short reactive utterances that 
mainly signal listening or prompt another participant to continue are annotated 
as backchannels rather than treated as independent turns. 
These include brief responses 
such as \textit{hai} (``yes''), \textit{ee} (``yes''), \textit{un} (``yeah/uh-huh''), 
and \textit{fuun} (``I see''), 
short repetitions of a prior utterance, 
and standalone interjections 
such as \textit{aa} (``ah'') or \textit{ee} (``oh'') 
when they are not followed by a more substantive utterance. 
The exception is that ``hai'' is treated as a turn when it functions as an answer to a question. 
The average number of turns per dialogue is 86.7, 
and the total number of turns is 3,121. 
Hereafter, for convenience, 
the three participants are referred to as A, B, and C.

\subsection{Addressee Annotation}

Based on \citet{kadota2024annotation}, 
each turn was assigned one of eight addressee labels.
Table~\ref{tab:addressee-label-defs} summarizes the label set used in this study.

\begin{table}[tb]
  \centering
  \small
  \begin{tabular}{ll}
    \toprule
    Label & Meaning \\
    \midrule
    \textit{A}, \textit{B}, \textit{C} & addressed to one specific participant \\
    \textit{E} & everyone \\
    \textit{W} & whoever \\
    \textit{N} & none \\
    \textit{O} & other \\
    \textit{U} & unknown \\
    \bottomrule
  \end{tabular}
  \caption{Addressee labels used in the annotation.}
  \label{tab:addressee-label-defs}
\end{table}

First, the authors conducted a pilot annotation on one dialogue. 
Based on this pilot, 
three annotators annotated the 3,078 turns in the remaining 35 dialogues. 
For each dialogue, 
three annotators performed the annotation independently. 
They were instructed to use multimodal information when judging the addressee, 
specifically explicit address terms, 
person references, 
register choice, 
references to another participant's prior experience or knowledge, 
links to prior talk such as disagreement markers (e.g., \textit{but}), 
deictic expressions and repetitions referring to previous utterances, 
whether the utterance was a response to a prior question, 
gaze direction, and pointing or other gestures. 
They were also instructed not to take into account any information 
occurring after the target turn.

\subsection{Gaze Annotation}

The gaze annotation classified each segment of a dialogue 
into one of three categories: 
the two participants other than the target participant (i.e., two of A, B, and C), 
or elsewhere (O). 
As with the addressee annotation, 
one dialogue was first annotated on a trial basis by students in our laboratory, 
and the remaining 35 dialogues were then annotated. 
Due to technical problems, 
some cameras stopped before the end of the dialogue in six dialogues; 
for those intervals after a camera stopped, 
no gaze annotation is available for the corresponding participant. 
In addition, 
intervals during which the gaze was shifting from one participant to another 
were classified as O, and participant labels were assigned only to intervals 
in which the annotator could judge that the gaze was clearly directed at that person.

\subsection{Preliminary Analysis}
\label{sec:preanalysis}

\begin{table}[tb]
  \centering
  \small
  \begin{tabular}{lr}
    \toprule
    Raw label & Count \\
    \midrule
    A & 1932 \\
    B & 1390 \\
    C & 1751 \\
    E & 3529 \\
    N & 247 \\
    O & 3 \\
    U & 48 \\
    W & 334 \\
    \bottomrule
  \end{tabular}
  \caption{Counts of raw addressee labels}
  \label{tab:raw-label-counts}
\end{table}

\begin{figure}[tb]
  \centering
  \includegraphics[width=.95\columnwidth]{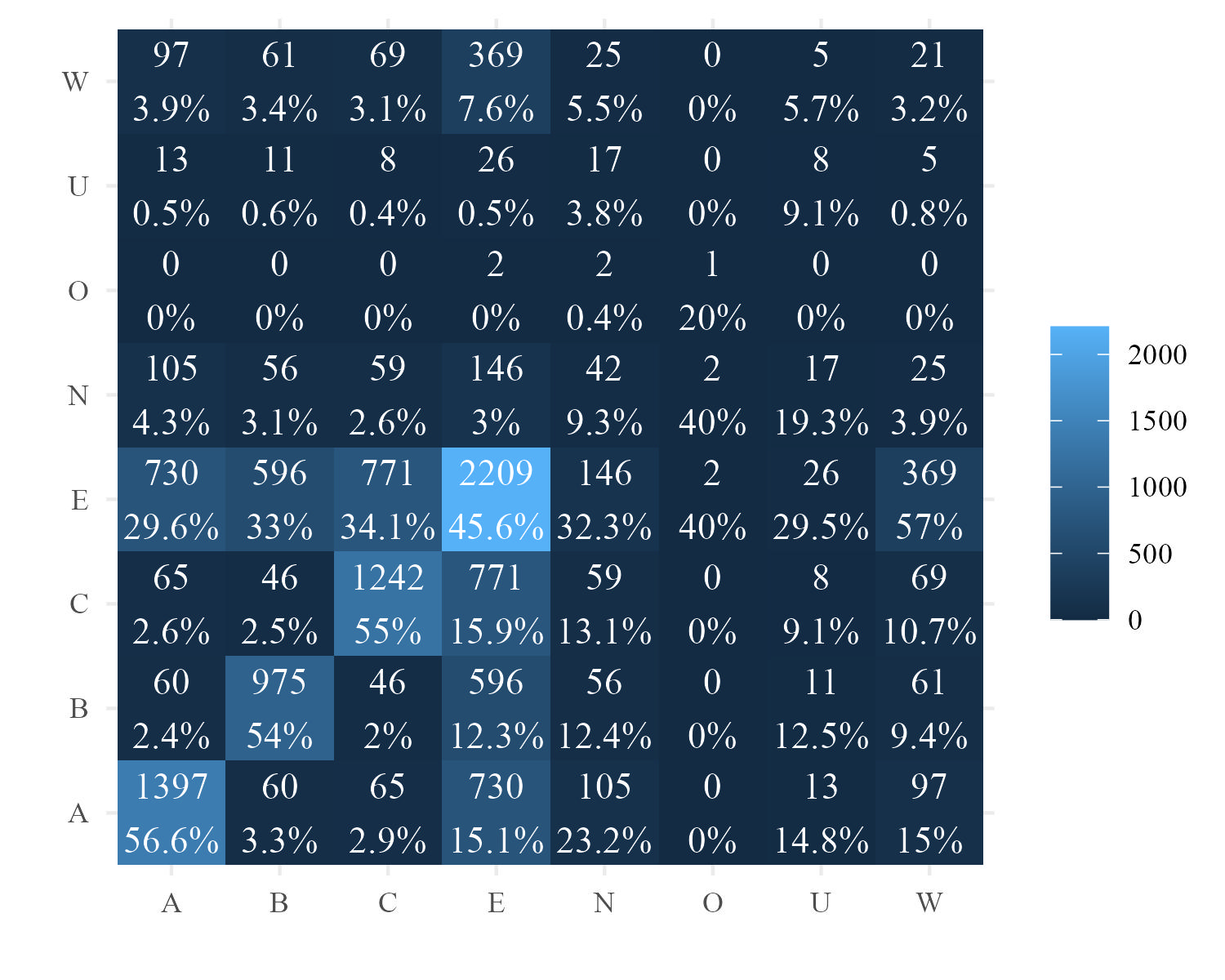}
  \caption{Co-occurrence heatmap of annotator address labels.}
  \label{fig:label-heatmap}
\end{figure}

To investigate addressee ambiguity, 
we analyzed the annotations for the 35 dialogues 
produced by the crowd workers. 
Table~\ref{tab:raw-label-counts} presents the frequency of each label. 
Among the participant-specific labels \textit{A}, \textit{B}, and \textit{C}, 
\textit{B} occurred somewhat less often than the other two, 
but all three appeared between 1,000 and 2,000 times. 
The most frequent label was \textit{E}, 
which occurred more than 3,500 times. 
By contrast, \textit{W}, \textit{N}, and \textit{O} were relatively infrequent. 
These results suggest that most utterances were judged 
to be clearly addressed either to a specific individual or to all participants. 
This is likely because the data consist of discussion-style dialogues, 
in which many utterances either question or challenge another participant's opinion or 
present the speaker's opinion to the group as a whole. 
As a measure of inter-annotator agreement, 
we calculated Fleiss' kappa. For the 3,078 turns annotated by three annotators, 
the agreement was $\kappa = 0.52$, indicating moderate agreement. 
This suggests that the addressee annotation was reasonably consistent overall, 
while still leaving room for ambiguity.

Next, to examine which labels were prone to ambiguity, 
Figure~\ref{fig:label-heatmap} shows a co-occurrence heatmap of label pairs. 
Co-occurrence counts were computed as follows: 
for example, if the three annotators labeled a turn as \textit{A}, \textit{A}, and \textit{B}, 
the pair \textit{AA} was counted once and the pair \textit{AB} was counted twice. 
The color intensity and the upper number in each cell indicate the raw count, 
while the lower number shows the proportion of that pair among all pairs involving the column label; 
these proportions sum to 100\% within each column.

For \textit{A}, \textit{B}, and \textit{C}, 
co-occurrence with the same label was the most frequent, 
exceeding 50\% in each case. 
The second most frequent co-occurrence was with \textit{E}, 
at around 30\%. 
A complementary pattern was observed for \textit{E}: 
co-occurrence with itself was the most frequent, 
at about 45\%, 
but co-occurrence with \textit{A}, \textit{B}, and \textit{C} was also substantial, 
at around 15\% each. 
These results indicate that 
judgments distinguishing utterances addressed to a specific individual 
from those addressed to the group were fairly stable, 
but not perfectly so. 
For \textit{N}, the most frequent co-occurrence was with \textit{E}, at 32\%. 
One possible reason is that both \textit{E} and \textit{N} share the property that 
the two other participants are treated similarly with respect to address. 
However, co-occurrence of \textit{N} with itself was only about 9\%, 
whereas co-occurrence with \textit{E} was about 3.5 times larger; 
given that \textit{E} itself occurred more than 14 times as often as \textit{N}, 
this does not appear to be excessive relative to chance. 
Like \textit{N}, \textit{W} most frequently co-occurred with \textit{E}, 
with a frequency about 17 times higher than its co-occurrence with itself. 
Since the difference in their marginal frequencies is only about tenfold, 
the co-occurrence between \textit{W} and \textit{E} is strikingly high 
even after taking label imbalance into account. 
This tendency was also noted in the prior study on which 
our label set was based \cite{kadota2024annotation}. 
Unlike \textit{N}, \textit{W} still implies that someone is being addressed, 
although not a specific individual, 
and in that respect it resembles \textit{E}. 
No meaningful pattern was found for \textit{O} or \textit{U} 
because they were too rare.

\section{Method}
\label{sec:method}
\subsection{Address Label and Address Level}

\begin{figure}[tb]
  \centering
  \includegraphics[width=.95\columnwidth]{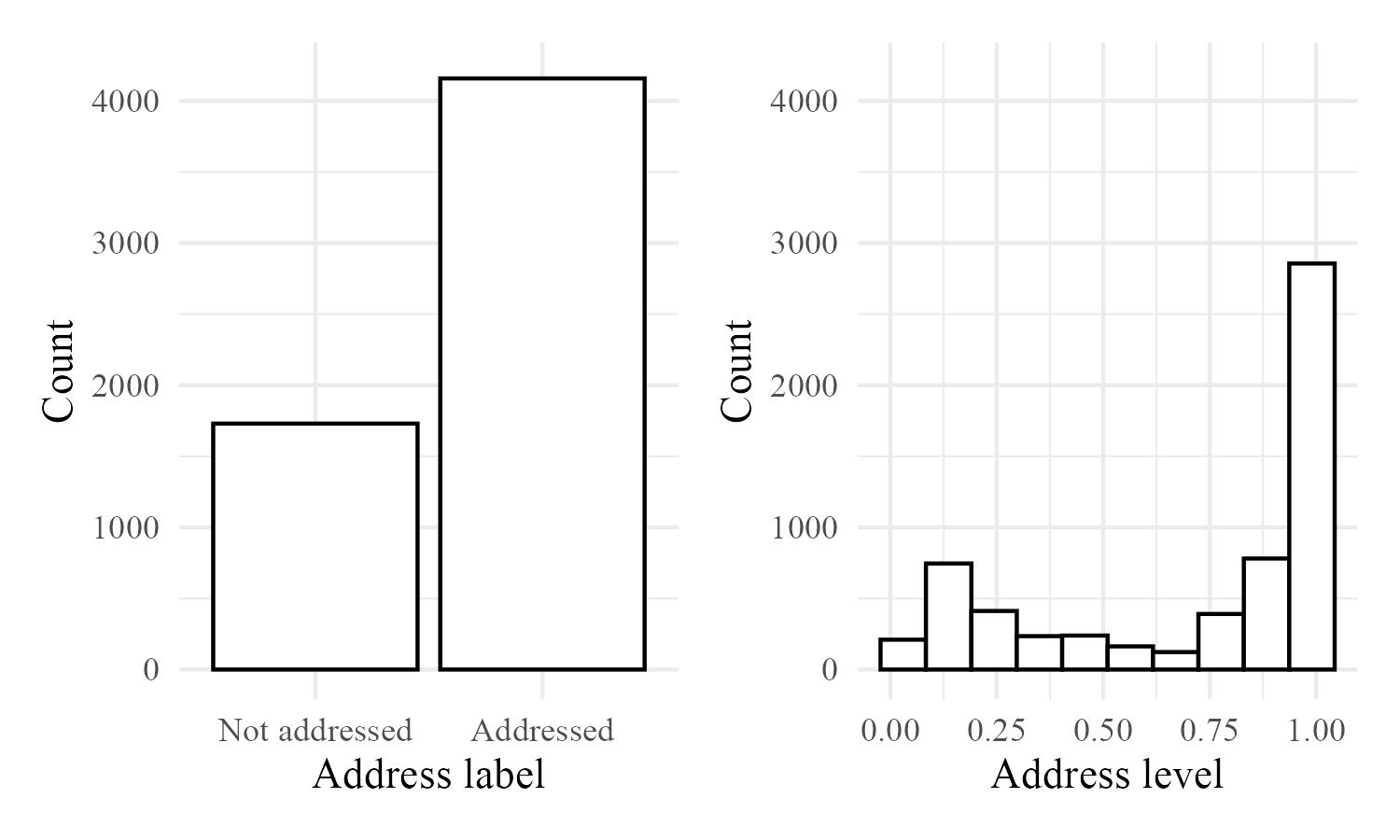}
  \caption{Distribution of address labels and address level.}
  \label{fig:address-label-level-count}
\end{figure}

Section~\ref{sec:preanalysis} showed that 
addressee annotation contains a certain degree of ambiguity. 
In this section, 
we argue that this ambiguity does not simply reflect annotator-specific errors, 
but may instead reflect ambiguity inherent in the act of addressing itself. 
We also examine whether 
address is related not only to turn-taking 
but also to other listener behaviors. 
To this end, we construct two representations of addressivity: 
an \textit{address label}, which follows the conventional discrete distinction 
between addressed and not addressed, and an \textit{address level}, 
which captures addressivity as a continuous property. 
We then compare these two representations.

We first describe how the address label was constructed. 
Based on the results in Section~\ref{sec:preanalysis}, 
\textit{W} was merged into \textit{E}. 
Next, the label \textit{U}, which indicates that 
the addressee could not be determined, 
was excluded. 
For the remaining six labels, 
a majority vote was taken for each turn, 
and when a majority label existed, 
it was used as the aggregated addressee label for that turn. 
Turns for which all annotators assigned \textit{U}, 
or for which no majority label was obtained, 
were excluded from this analysis. 
In our data, 
the former case did not occur, 
whereas 134 turns fell into the latter category 
and were excluded. 
Then, for each of the two listeners in each turn, 
an \textit{addressed} label was assigned 
if the majority addressee matched that listener (i.e., \textit{A}, \textit{B}, or \textit{C}) 
or included that listener (i.e., \textit{E}). 
Conversely, a \textit{not addressed} label was assigned 
if the majority addressee matched the other listener, 
or if the utterance was labeled as monologic speech (\textit{N}) 
or as being addressed to an entity outside the participants (\textit{O}). 
This binary value was used as the address label for each listener in each turn.

Next, 
we describe how the address level was estimated. 
Unlike the address label, 
the address level was treated as a latent continuous variable 
inferred from the labels assigned by the annotators. 
For each turn \(i\), 
we assumed two latent addressivity values, 
\(\theta_{i1}\) and \(\theta_{i2}\), 
corresponding to the two listeners of the current speaker. 
These values represent the degree to which the utterance is addressed 
to listener 1 and listener 2, respectively. 
They were constrained to the unit interval by applying a logistic transformation 
to normally distributed latent scores:
\begin{linenomath*}
\begin{align}
\theta_{i1} &= \mathrm{logit}^{-1}(\eta_{i1}), \\
\theta_{i2} &= \mathrm{logit}^{-1}(\eta_{i2}).
\end{align}
\end{linenomath*}

The annotator labels were modeled as categorical observations 
generated from the latent addressivity vector 
\(\boldsymbol{\theta}_i = (\theta_{i1}, \theta_{i2})\). 
As with the address label, 
\textit{W} was merged into \textit{E}, 
labels referring to the current speaker and \textit{O} were merged into \textit{N}, 
and \textit{U} was excluded from the likelihood. 
The remaining labels were associated with ideal points in the two-dimensional addressivity space: 
\textit{L1} with \((1,0)\), 
\textit{L2} with \((0,1)\), 
\textit{E} with \((1,1)\), 
and \textit{N} with \((0,0)\). 
Here, \textit{L1} and \textit{L2} denote the two listeners of the current speaker. 
We treat addressivity as listener-wise directedness 
rather than as a finite resource to be divided among listeners; 
therefore, \textit{E} was represented as \((1,1)\) 
rather than as a normalized allocation such as \((0.5,0.5)\).
For annotator \(j\) and label \(c\), 
the probability of the observed label \(y_{ij}\) was defined as
\begin{linenomath*}
\begin{equation}
\Pr(y_{ij}=c)
=
\mathrm{softmax}_c
\left[
\alpha_{jc}
-
\lambda_c
\left\|
\boldsymbol{\theta}_i - \mathbf{v}_c
\right\|^2
\right],
\end{equation}
\end{linenomath*}
where \(\mathbf{v}_c\) is the ideal point for label \(c\), 
\(\alpha_{jc}\) is an annotator-specific label bias, 
and \(\lambda_c\) is a label-specific discrimination parameter. 
Thus, labels whose ideal points were closer to the latent addressivity vector 
were assigned higher probabilities, 
while allowing annotators to differ in their tendencies to use particular labels.
Although annotators may also differ in label-specific discrimination, 
the discrimination parameters were shared across annotators 
to avoid overparameterization, 
given that each turn was annotated by only three annotators. 
Annotator-level differences in label-use tendencies were instead captured 
by the bias parameters \(\alpha_{jc}\).

The latent-address model was estimated by Bayesian sampling 
using R version 4.6.0 and \texttt{cmdstanr} version 0.9.0. 
We used four chains, 
with 2,000 sampling iterations per chain 
and 1,000 warm-up iterations. 
For each turn-listener pair, 
the posterior median of the corresponding latent addressivity value 
was used as the address level in the subsequent analyses.

Figure~\ref{fig:address-label-level-count} shows the distributions of the address labels and address levels. 
For the address labels, 
the number of \textit{addressed} instances is approximately twice that of \textit{not addressed}. 
This is consistent with the original label distribution, 
in which \textit{E} was the most frequent label, 
followed by the participant-specific labels (\textit{A}, \textit{B}, and \textit{C}). 
For the address levels, 
many observations are concentrated near the high ends of the scale, 
but intermediate values are also observed, 
indicating the existence of partially addressed states.

\FloatBarrier

\subsection{Modeling}

Next, we analyzed the relationships between these two address measures 
and turn-taking, listener gaze, and backchannel behavior. 
Each sample represented the behavior of each listener in each turn, 
yielding hierarchical data nested within turns, 
dialogues, speakers, and listeners. 
Accordingly, 
we modeled the data using generalized linear mixed-effects models 
and estimated the parameters by Bayesian sampling.
The mixed-effects models were fitted using the \texttt{brms} package version 2.23.0. 
For all models, 
we used four chains, 
with 2,000 sampling iterations per chain 
and 1,000 warm-up iterations. 
As weakly informative priors, 
parameters defined over the entire real line, 
such as intercepts and regression coefficients, 
were assigned normal distributions with mean 0 and standard deviation 10, 
whereas non-negative parameters, 
such as sd and sigma, were assigned half-Cauchy distributions with location 0 and scale 10. 
In all models, the fixed effect was either address label or address level, 
and the random effects were random intercepts for turn, dialogue session, speaker, and listener. 
These random intercepts were defined as follows:
\begin{linenomath*}
\begin{align*}
u_{turn,i} &\sim \mathcal{N}(0, \sigma_{\mathrm{turn}}) \\
u_{session,i} &\sim \mathcal{N}(0, \sigma_{\mathrm{session}}) \\
u_{speaker,i} &\sim \mathcal{N}(0, \sigma_{\mathrm{speaker}}) \\
u_{listener,i} &\sim \mathcal{N}(0, \sigma_{\mathrm{listener}})
\end{align*}
\end{linenomath*}
Convergence was assessed 
using diagnostics such as $\hat{R}$, Bulk ESS, and trace plots, 
and model comparisons were conducted 
using the Expected Log Predictive Density (ELPD) estimated with the \texttt{loo} function.
ELPD evaluates out-of-sample predictive performance, with larger values indicating better expected predictive fit. 
To make the ELPD comparisons between the address-label and address-level models valid, 
the two models for each outcome were fit to the same set of observations. 
Therefore, turns for which no majority-vote address label was obtained 
were also excluded from the address-level models.

Turn-taking was modeled as follows.
\begin{linenomath*}
\begin{equation*}
y_i \sim \mathrm{Bernoulli}(p_i)
\end{equation*}
\begin{subequations}
\begin{equation}
\label{f_next_lab}
\begin{split}
\mathrm{logit}(p_i) &= \beta_0 + \beta_1\,x_{\mathrm{address\_label},i} + u_{turn,i} \\
&\quad + u_{session,i} + u_{speaker,i} + u_{listener,i}
\end{split}
\end{equation}
\begin{equation}
\label{f_next_lev}
\begin{split}
\mathrm{logit}(p_i) &= \beta_0 + \beta_1\,x_{\mathrm{address\_level},i} + u_{turn,i} \\
&\quad + u_{session,i} + u_{speaker,i} + u_{listener,i}
\end{split}
\end{equation}
\end{subequations}
\end{linenomath*}

\noindent Here, \(y_i\) is a binary variable indicating 
whether the focal listener in the turn corresponding to the \(i\)-th observation 
became the next speaker, 
and it is assumed to follow a Bernoulli distribution with parameter \(p_i\). 
The predictors \(x_{\mathrm{address\_label},i}\) and \(x_{\mathrm{address\_level},i}\) 
denote the address label and address level for the focal listener, respectively. 
The random intercepts \(u_{turn,i}\), \(u_{session,i}\), \(u_{speaker,i}\), and \(u_{listener,i}\) 
represent variation associated with the turn, dialogue session, current speaker, and focal listener. 
The total number of observations was 5,820, 
calculated as the number of turns 
multiplied by the two listeners, 
excluding turns near the end of a dialogue 
for which no next speaker existed.

Listener gaze was modeled as follows.
\begin{linenomath*}
\begin{equation*}
y_i \sim \mathrm{Beta}(\mu_i, \phi)
\end{equation*}
\begin{subequations}
\begin{equation}
\label{f_gaze_lab}
\begin{split}
\mathrm{logit}(\mu_i) &= \beta_0 + \beta_1\,x_{\mathrm{address\_label},i} + u_{turn,i} \\
&\quad + u_{session,i} + u_{speaker,i} + u_{listener,i}
\end{split}
\end{equation}
\begin{equation}
\label{f_gaze_lev}
\begin{split}
\mathrm{logit}(\mu_i) &= \beta_0 + \beta_1\,x_{\mathrm{address\_level},i} + u_{turn,i} \\
&\quad + u_{session,i} + u_{speaker,i} + u_{listener,i}
\end{split}
\end{equation}
\end{subequations}
\end{linenomath*}

\noindent Here, \(y_i\) is the proportion of time 
during the turn corresponding to the \(i\)-th observation 
for which the focal listener was looking at the current speaker, 
and it is assumed to follow a Beta distribution with mean parameter \(\mu_i\). 
This distribution was used because the outcome is a proportion bounded between 0 and 1. 
Because the data include values of 0 and 1, 
the observed proportions were adjusted using the method of \citet{smithson2006better} 
before fitting the Beta model. 
Specifically, this transformation shrinks the observed proportions 
slightly toward the interior of the unit interval, 
so that values of 0 and 1 are mapped to values within $(0,1)$ 
while preserving their relative ordering.
The total number of observations was 5,744, 
excluding intervals 
for which gaze annotation was unavailable due to technical problems.

Backchannel was modeled as follows.

\begin{linenomath*}
\begin{equation*}
y_i \sim \mathrm{Poisson}(\mu_i)
\end{equation*}
\begin{subequations}
\begin{equation}
\label{f_bc_lab}
\begin{split}
\log(\mu_i) &= \beta_0 + \beta_1\,x_{\mathrm{address\_label},i} \\
&\quad + \log(\mathrm{duration}_i) + u_{turn,i} \\
&\quad + u_{session,i} + u_{speaker,i} + u_{listener,i}
\end{split}
\end{equation}
\begin{equation}
\label{f_bc_lev}
\begin{split}
\log(\mu_i) &= \beta_0 + \beta_1\,x_{\mathrm{address\_level},i} \\
&\quad + \log(\mathrm{duration}_i) + u_{turn,i} \\
&\quad + u_{session,i} + u_{speaker,i} + u_{listener,i}
\end{split}
\end{equation}
\end{subequations}
\end{linenomath*}

\noindent Here, \(y_i\) is the number of backchannels 
produced by the focal listener 
during the turn corresponding to the \(i\)-th observation, 
and it is assumed to follow a Poisson distribution. 
This distribution was used because the outcome is a count variable. 
The expected count is modeled as a function of fixed and random effects, 
with the duration of the turn included as an offset term. 
Accordingly, 
the model effectively estimates the rate of backchannel production 
rather than the raw count alone. 
Very short turns were excluded from the analysis 
because they often consisted of incomplete utterances 
or similarly brief segments 
for which backchanneling is unlikely in principle. 
Consistent with this, 
more than 90\% of turns shorter than one second had zero backchannels. 
After excluding these short turns, 
the total number of observations used for the backchannel analysis 
was 4,920.

\section{Results}
\subsection{Next Speaker}

\begin{figure}[tb]
  \centering
  \includegraphics[width=\columnwidth]{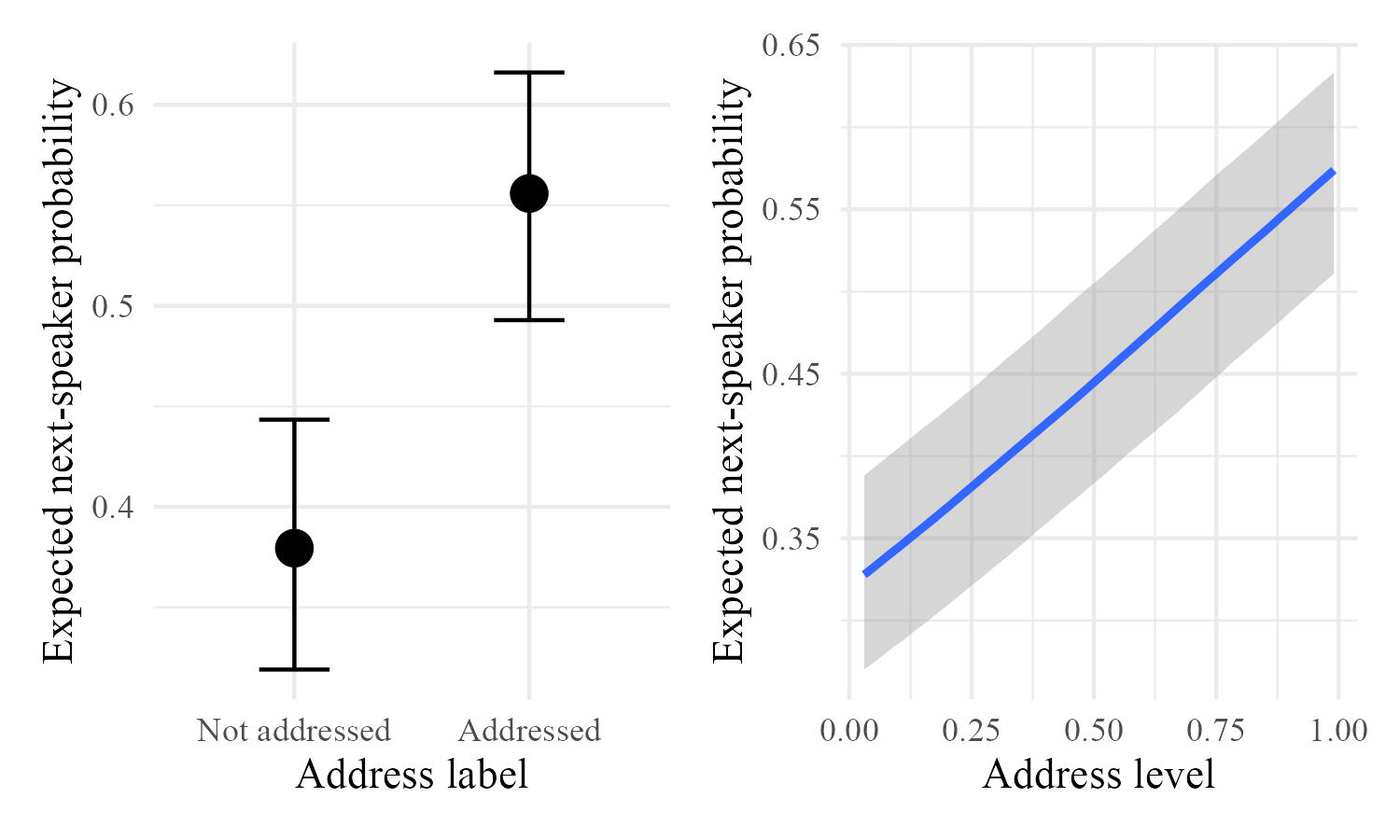}
  \caption{Posterior conditional effects of addressivity on next-speaker selection.}
  \label{fig:next-speaker}
\end{figure}

Figure~\ref{fig:next-speaker} shows the posterior conditional effects of addressivity 
in Models 1a and 1b. 
Full posterior summaries of the fixed and random effects are reported in Appendix~\ref{sec:results-next}. 
As shown in the figure, 
both models indicate that 
listeners with higher addressivity are more likely to become the next speaker. 
In Model 1a, the coefficient for \textit{addressed} was positive 
(Est. = 0.72, 95\% CI [0.59, 0.84]), 
and in Model 1b, the coefficient for \textit{addressLevel} was also positive 
(Est. = 1.06, 95\% CI [0.89, 1.23]), 
indicating clear positive effects of addressivity on next-speaker selection. 
The 95\% credible intervals for both coefficients excluded zero. 
Table~\ref{tab:next-loo} reports the leave-one-out comparison. 
Model 1b, which uses address level, 
provides better predictive fit than Model 1a. 
This suggests that next-speaker selection is better captured 
by a continuous representation of addressivity than 
by a discrete address label alone. 
In other words, listeners who would be categorized as not addressed under a discrete formulation 
may still become the next speaker, 
whereas listeners categorized as addressed do not always do so.

\begin{table}[tb]
  \centering
  \small
  \begin{tabular}{lrr}
    \toprule
    Model & elpd\_diff & se\_diff \\
    \midrule
    (1b) & 0.0 & 0.0 \\
    (1a)* & -14.4 & 4.9 \\
    \bottomrule
  \end{tabular}
  \caption{Leave-one-out comparison for the next-speaker models. Values are shown relative to the best model. An asterisk (*) indicates a substantial difference in predictive fit, defined here as $|\mathrm{elpd\_diff}| > 1.96 \times \mathrm{se\_diff}$.}
  \label{tab:next-loo}
\end{table}

\subsection{Gaze}

\begin{figure}[tb]
  \centering
  \includegraphics[width=\columnwidth]{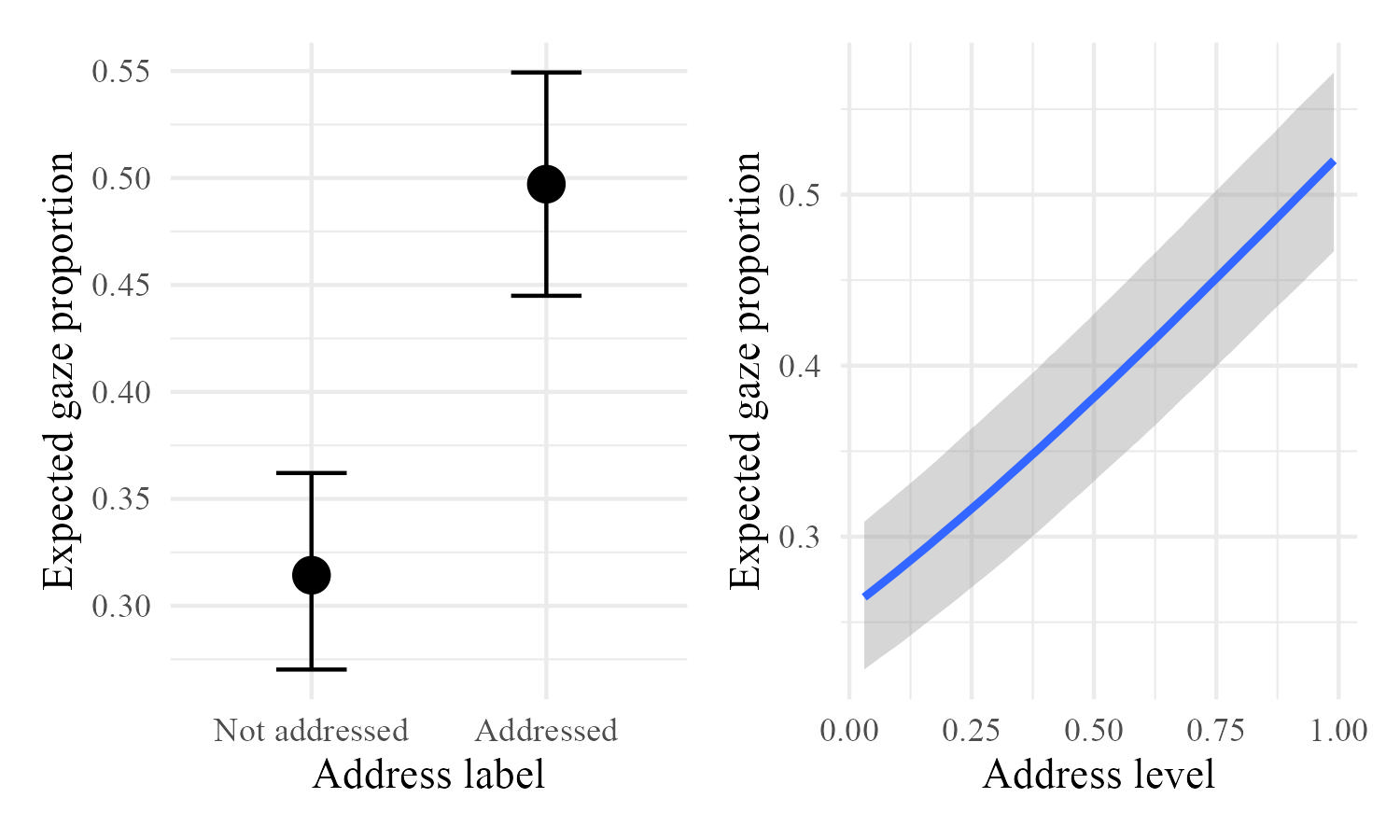}
  \caption{Posterior conditional effects of addressivity on listener gaze toward the current speaker.}
  \label{fig:gaze}
\end{figure}

Figure~\ref{fig:gaze} shows the posterior conditional effects of addressivity 
in Models 2a and 2b. 
Full posterior summaries of the fixed and random effects are reported in Appendix~\ref{sec:results-gaze}. 
As shown in the figure, 
both models indicate that 
listeners with higher addressivity show higher gaze proportions toward the current speaker. 
In Model 2a, the coefficient for \textit{addressed} was positive 
(Est. = 0.77, 95\% CI [0.69, 0.85]), 
and in Model 2b, the coefficient for \textit{addressLevel} was also positive 
(Est. = 1.15, 95\% CI [1.04, 1.26]), 
indicating clear positive effects of addressivity on listener gaze.  
Table~\ref{tab:gaze-loo} reports the leave-one-out comparison. 
Model 2b, which uses address level, 
provides better predictive fit than Model 2a. 
This suggests that listeners with higher addressivity tend to look at the speaker 
for longer periods of time,
and that listener gaze is better captured 
by a continuous representation of addressivity than 
by a discrete address label alone.
In other words, gaze toward the current speaker varies in a graded manner 
with the strength of addressivity, 
rather than being fully explained by a binary addressed/not-addressed distinction.

\begin{table}[tb]
  \centering
  \small
  \begin{tabular}{lrr}
    \toprule
    Model & elpd\_diff & se\_diff \\
    \midrule
    (2b) & 0.0 & 0.0 \\
    (2a)* & -39.2 & 7.9 \\
    \bottomrule
  \end{tabular}
  \caption{Leave-one-out comparison for the listener-gaze models. Values are shown relative to the best model.}
  \label{tab:gaze-loo}
\end{table}

\subsection{Backchannel}

\begin{figure}[tb]
  \centering
  \includegraphics[width=\columnwidth]{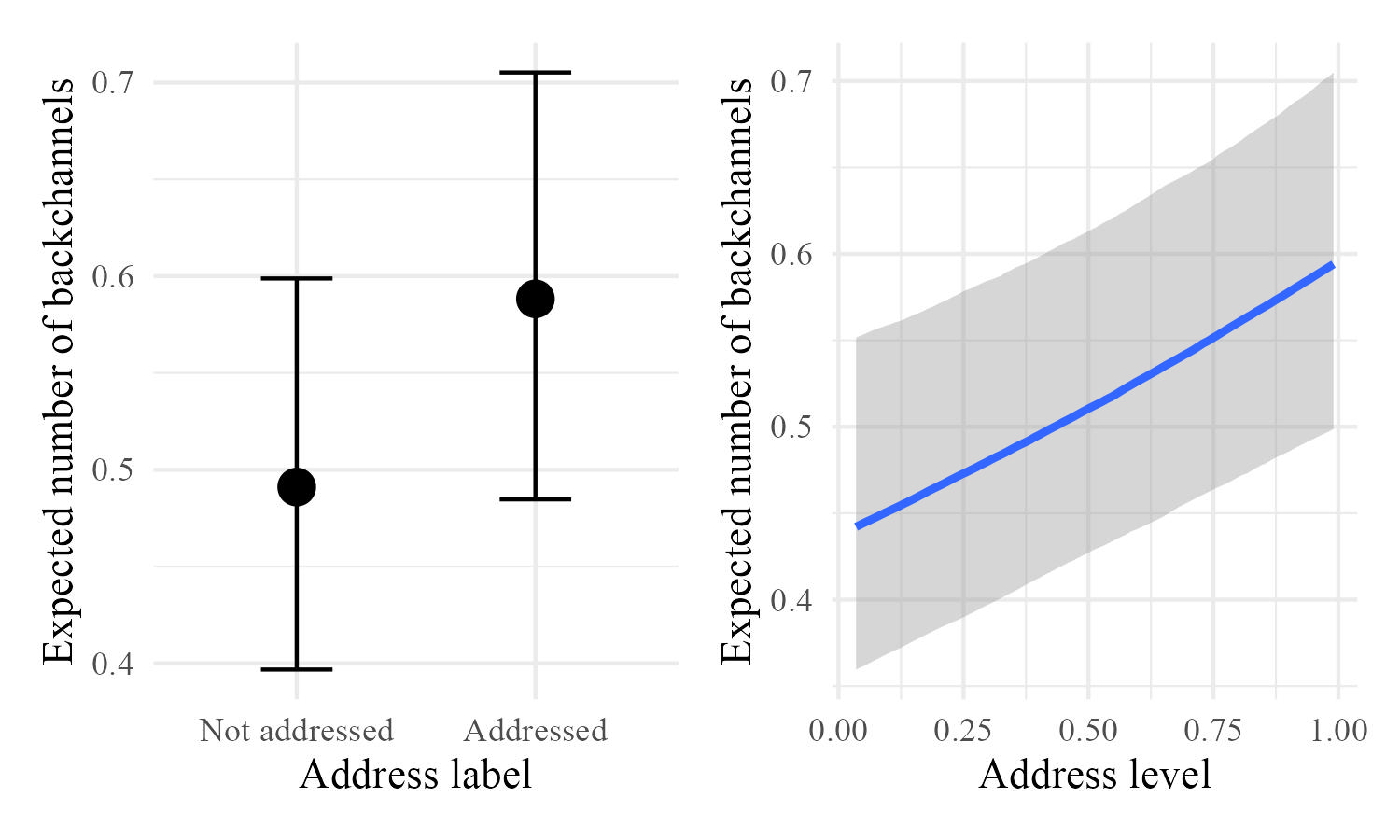}
  \caption{Posterior conditional effects of addressivity on listener backchannel production.}
  \label{fig:backchannel}
\end{figure}

Figure~\ref{fig:backchannel} shows the posterior conditional effects of addressivity 
in Models 3a and 3b. 
Full posterior summaries of the fixed and random effects are reported in Appendix~\ref{sec:results-bc}. 
As shown in the figure, 
both models indicate that 
listeners with higher addressivity tend to produce more backchannels. 
In Model 3a, the coefficient for \textit{addressed} was positive 
(Est. = 0.18, 95\% CI [0.08, 0.28]), 
whereas in Model 3b, the coefficient for \textit{addressLevel} was also positive 
(Est. = 0.31, 95\% CI [0.17, 0.45]), 
indicating clear effects of addressivity on backchannel production. 
The 95\% credible intervals for both coefficients excluded zero. 
Table~\ref{tab:bc-loo} reports the leave-one-out comparison. 
Model 3b, which uses address level, 
provides better predictive fit than Model 3a. 
This suggests that listeners with higher addressivity tend to produce more backchannels,
and that backchannel production is better captured 
by a continuous representation of addressivity than 
by a discrete address label alone.
However, at the same time, 
the improvement is smaller than for next-speaker selection and listener gaze, 
suggesting that backchanneling is influenced not only by addressivity 
but also by additional factors.

\begin{table}[tb]
  \centering
  \small
  \begin{tabular}{lrr}
    \toprule
    Model & elpd\_diff & se\_diff \\
    \midrule
    (3b) & 0.0 & 0.0 \\
    (3a)* & -3.5 & 1.6 \\
    \bottomrule
  \end{tabular}
  \caption{Leave-one-out comparison for the backchannel models. Values are shown relative to the best model.}
  \label{tab:bc-loo}
\end{table}

\FloatBarrier

\subsection{Qualitative Analysis}
In the following excerpt, the participants are discussing 
what means of transportation they would use if their club were to travel to Tokyo. 
Due to space limitations, we present only an English translation 
that is as faithful as possible to the original Japanese.

In line 01, C asks what they are supposed to do in Tokyo. 
While asking this question, C gazes at B, thereby strongly addressing B. 
Indeed, the next speaker in line 02 is B. 
Since line 01 is a question, that is, the first pair part of an adjacency pair, 
an answer as the second pair part is conditionally relevant in the next turn. 
However, what B produces in line 02 is another question. 
This question is not unrelated to C's question in line 01. 
While gazing at A, 
B asks whether it had been decided in the first place what they would do in Tokyo. 
In other words, B questions the presupposition of C's question. 
A then answers ``shopping'' in line 03.
Interestingly, although A's answer is sequentially occasioned by B's question, 
A gazes at C and, in terms of content, answers C's question. 
For this reason, two annotators assigned the label C to this utterance. 
However, when observing the interaction, 
we do not get the impression that B's question is being ignored. 
This is because B's question concerns the presupposition of C's question, 
and answering C's question also provides an affirmative answer to B's question. 
In other words, A's utterance functions as an answer, 
or second pair part, not only to C's question but also to B's question. 
Thus, although the primary address is directed toward C, 
there is also a weaker degree of address toward B. 
Presumably for this reason, the remaining annotator assigned the label E.

{
\setlength{\parindent}{0pt}
Excerpt1

01 C: Wait, what were we supposed to do in Tokyo again?

02 B: Was there something we were supposed to do in Tokyo?

03 A: Shopping.

}

\section{Conclusions}

The findings of this study have several implications for future research on 
addressee detection.
First, the conventional framework of aggregating annotations 
into a single label and formulating the task 
as multiclass classification may oversimplify the act of addressing 
and discard important aspects of the phenomenon. 
In contrast, 
preserving variation across annotators and representing address 
continuously may provide a more appropriate way of modeling the phenomenon. 
The advantages of modeling address continuously are not limited to descriptive adequacy, 
but are also practically relevant for dialogue systems. 
For example, in conventional multiclass classification, 
utterances addressed to the whole group are collapsed into a single label, 
which by itself may not provide sufficient information for deciding turn allocation. 
By contrast, a continuous representation of address can capture subtle differences 
across participants even within cases that would previously have been treated uniformly as group-addressed. 
Moreover, 
a framework that estimates participant-wise address levels is less dependent on the number of participants 
in a dialogue and may therefore support more scalable model designs.

Second, the finding that address is related to listener behaviors 
during the speaker's turn suggests that 
address prediction should not be treated solely as a problem at turn completion, 
but rather as something that should be inferred online during ongoing speech. 
In the future, 
it will be important to develop frameworks that estimate addressivity in real time 
while jointly modeling listener behaviors such as gaze and backchannels, 
as well as turn-taking.

Reconsidered in light of the turn-taking model of \citet{sacks1974simplest}, 
the present results suggest that the current speaker may be understood 
not as discretely selecting one next speaker or addressee, 
but as continuously distributing degrees of addressivity across participants 
through multimodal interactional resources \citep{kadota2024annotation}, 
thereby using listener-wise distributions of addressivity to fine-tune 
both next-speaker relevance and the uptake of the ongoing action.

Future work should examine whether these findings generalize across languages, 
domains, and group sizes, 
and should develop more refined methods for constructing and validating continuous representations of addressivity. 
Given the links observed here between addressivity and listener gaze and backchannels, 
future systems may benefit from estimating addressivity online 
while jointly modeling multiple listener behaviors.

\section*{Acknowledgments}

This work was supported by JST Moonshot R\&D JPMJPS2011 and JST PRESTO JPMJPR24I4.

\section*{Limitations}
The data are limited to Japanese triadic discussions, 
and gaze behavior and backchannel production may vary across languages, cultures, settings, and participant characteristics. 
Although address level was estimated as a latent continuous variable, 
it was inferred from only three categorical annotations per turn; 
the behavioral analyses also used posterior medians rather than propagating full posterior uncertainty. 
Finally, annotators were allowed to use gaze information when judging addressees. 
Because this primarily concerned the speaker's gaze, whereas our analysis focused on listener gaze toward the current speaker, 
the gaze analysis does not simply reuse the same behavioral signal, 
but it should still be interpreted as evidence of consistency rather than as fully independent validation.

\bibliography{custom}

\appendix
\section{Detailed Results for the Next-Speaker Models}
\label{sec:results-next}

This appendix reports the full posterior summaries for Models 1a and 1b for next-speaker selection. 
Both models were fit to 5,820 observations using four chains with 2,000 iterations per chain, including 1,000 warm-up iterations, yielding 4,000 post-warm-up draws in total.

Table~\ref{tab:app-next-fixed} summarizes the fixed effects. In Model 1a, the coefficient for \textit{addressed} is positive, indicating that listeners categorized as addressed were more likely to become the next speaker than non-addressed listeners. In Model 1b, the coefficient for \textit{addressLevel} is positive, indicating that the probability of next-speaker selection increases as address level increases.

\begin{table}[t]
  \centering
  \scriptsize
  \begin{tabular}{llrrrr}
    \toprule
    Model & Param. & Est. & SE & l-95\% CI & u-95\% CI \\
    \midrule
    (1a) & Intercept* & -0.49 & 0.13 & -0.76 & -0.23 \\
    (1a) & addressed* & 0.72 & 0.06 & 0.59 & 0.84 \\
    (1b) & Intercept* & -0.75 & 0.14 & -1.03 & -0.49 \\
    (1b) & addressLevel* & 1.06 & 0.09 & 0.89 & 1.23 \\
    \bottomrule
  \end{tabular}
  \caption{Posterior summaries of the fixed effects for the next-speaker models. An asterisk (*) indicates that the 95\% credible interval does not include zero.}
  \label{tab:app-next-fixed}
\end{table}

Table~\ref{tab:app-next-random} reports the posterior summaries of the random-effect standard deviations. In both models, listener-level variation was the largest source of heterogeneity, followed by speaker-level variation, whereas session- and turn-level variation were comparatively small.

\begin{table}[t]
  \centering
  \scriptsize
  \begin{tabular}{llrrrr}
    \toprule
    Model & Random effect & Est. & SE & l-95\% CI & u-95\% CI \\
    \midrule
    (1a) & listenerID & 0.65 & 0.09 & 0.49 & 0.84 \\
    (1a) & sessionID & 0.03 & 0.02 & 0.00 & 0.09 \\
    (1a) & speakerID & 0.30 & 0.06 & 0.19 & 0.44 \\
    (1a) & turnID & 0.02 & 0.02 & 0.00 & 0.06 \\
    (1b) & listenerID & 0.64 & 0.09 & 0.49 & 0.85 \\
    (1b) & sessionID & 0.03 & 0.02 & 0.00 & 0.08 \\
    (1b) & speakerID & 0.29 & 0.06 & 0.18 & 0.43 \\
    (1b) & turnID & 0.02 & 0.02 & 0.00 & 0.06 \\
    \bottomrule
  \end{tabular}
  \caption{Posterior summaries of the random-effect standard deviations for the next-speaker models.}
  \label{tab:app-next-random}
\end{table}

All parameters showed satisfactory convergence diagnostics, with $\hat{R} = 1.00$ throughout. Bulk ESS and Tail ESS were also sufficiently large for all reported parameters.

\section{Detailed Results for the Listener-Gaze Models}
\label{sec:results-gaze}

This appendix reports the full posterior summaries for Models 2a and 2b for listener gaze. Both models were fit to 5,744 observations using four chains with 2,000 iterations per chain, including 1,000 warm-up iterations, yielding 4,000 post-warm-up draws in total.

Table~\ref{tab:app-gaze-fixed} summarizes the fixed effects and the beta precision parameter. In Model 2a, the coefficient for \textit{addressed} is positive, indicating that listeners categorized as addressed showed higher gaze proportions toward the current speaker. In Model 2b, the coefficient for \textit{addressLevel} is also positive, indicating that expected gaze proportion increases as address level increases.

\begin{table}[t]
  \centering
  \scriptsize
  \begin{tabular}{llrrrr}
    \toprule
    Model & Param. & Est. & SE & l-95\% CI & u-95\% CI \\
    \midrule
    (2a) & Intercept* & -0.78 & 0.11 & -0.99 & -0.57 \\
    (2a) & addressed* & 0.77 & 0.04 & 0.69 & 0.85 \\
    (2a) & $\phi$* & 0.83 & 0.02 & 0.79 & 0.88 \\
    (2b) & Intercept* & -1.06 & 0.11 & -1.29 & -0.84 \\
    (2b) & addressLevel* & 1.15 & 0.05 & 1.04 & 1.26 \\
    (2b) & $\phi$* & 0.84 & 0.02 & 0.79 & 0.89 \\
    \bottomrule
  \end{tabular}
  \caption{Posterior summaries of the fixed effects and beta precision parameter for the listener-gaze models. An asterisk (*) indicates that the 95\% credible interval does not include zero.}
  \label{tab:app-gaze-fixed}
\end{table}

Table~\ref{tab:app-gaze-random} reports the posterior summaries of the random-effect standard deviations. In both models, turn-level variation was the largest source of heterogeneity, followed by listener- and speaker-level variation, whereas session-level variation was comparatively smaller.

\begin{table}[t]
  \centering
  \scriptsize
  \begin{tabular}{llrrrr}
    \toprule
    Model & Random effect & Est. & SE & l-95\% CI & u-95\% CI \\
    \midrule
    (2a) & listenerID & 0.47 & 0.07 & 0.35 & 0.63 \\
    (2a) & sessionID & 0.22 & 0.05 & 0.14 & 0.32 \\
    (2a) & speakerID & 0.35 & 0.06 & 0.25 & 0.49 \\
    (2a) & turnID & 0.58 & 0.04 & 0.50 & 0.65 \\
    (2b) & listenerID & 0.47 & 0.07 & 0.35 & 0.63 \\
    (2b) & sessionID & 0.21 & 0.04 & 0.14 & 0.31 \\
    (2b) & speakerID & 0.34 & 0.06 & 0.24 & 0.48 \\
    (2b) & turnID & 0.57 & 0.04 & 0.49 & 0.64 \\
    \bottomrule
  \end{tabular}
  \caption{Posterior summaries of the random-effect standard deviations for the listener-gaze models.}
  \label{tab:app-gaze-random}
\end{table}

Convergence diagnostics were satisfactory overall. Most reported parameters had $\hat{R}$ values of 1.00, while the turn-level standard deviation and $\phi$ in Model 2b had $\hat{R} = 1.01$, which still indicates acceptable convergence. Bulk ESS and Tail ESS were also sufficiently large for the reported parameters.

\section{Detailed Results for the Backchannel Models}
\label{sec:results-bc}

This appendix reports the full posterior summaries for Models 3a and 3b for backchannel production. Both models were fit to 4,920 observations using four chains with 2,000 iterations per chain, including 1,000 warm-up iterations, yielding 4,000 post-warm-up draws in total.

Table~\ref{tab:app-bc-fixed} summarizes the fixed effects. In Model 3a, the coefficient for \textit{addressed} is positive, indicating that listeners categorized as addressed produced more backchannels than non-addressed listeners. In Model 3b, the coefficient for \textit{addressLevel} is positive, indicating that the expected backchannel rate increases as address level increases.

\begin{table}[t]
  \centering
  \scriptsize
  \begin{tabular}{llrrrr}
    \toprule
    Model & Param. & Est. & SE & l-95\% CI & u-95\% CI \\
    \midrule
    (3a) & Intercept* & -2.37 & 0.10 & -2.58 & -2.17 \\
    (3a) & addressed* & 0.18 & 0.05 & 0.08 & 0.28 \\
    (3b) & Intercept* & -2.48 & 0.11 & -2.69 & -2.26 \\
    (3b) & addressLevel* & 0.31 & 0.07 & 0.17 & 0.45 \\
    \bottomrule
  \end{tabular}
  \caption{Posterior summaries of the fixed effects for the backchannel models. An asterisk (*) indicates that the 95\% credible interval does not include zero.}
  \label{tab:app-bc-fixed}
\end{table}

Table~\ref{tab:app-bc-random} reports the posterior summaries of the random-effect standard deviations. In both models, listener-level variation was the largest source of heterogeneity, whereas session-, speaker-, and turn-level variation were comparatively small.

\begin{table}[t]
  \centering
  \scriptsize
  \begin{tabular}{llrrrr}
    \toprule
    Model & Random effect & Est. & SE & l-95\% CI & u-95\% CI \\
    \midrule
    (3a) & listenerID & 0.51 & 0.07 & 0.38 & 0.67 \\
    (3a) & sessionID & 0.09 & 0.04 & 0.01 & 0.18 \\
    (3a) & speakerID & 0.12 & 0.05 & 0.03 & 0.22 \\
    (3a) & turnID & 0.05 & 0.03 & 0.00 & 0.12 \\
    (3b) & listenerID & 0.50 & 0.07 & 0.38 & 0.66 \\
    (3b) & sessionID & 0.08 & 0.04 & 0.01 & 0.17 \\
    (3b) & speakerID & 0.12 & 0.05 & 0.02 & 0.23 \\
    (3b) & turnID & 0.04 & 0.03 & 0.00 & 0.12 \\
    \bottomrule
  \end{tabular}
  \caption{Posterior summaries of the random-effect standard deviations for the backchannel models.}
  \label{tab:app-bc-random}
\end{table}

Convergence diagnostics were satisfactory overall. Most reported parameters had $\hat{R}$ values of 1.00, while a small number of parameters had $\hat{R} = 1.01$, which still indicates acceptable convergence. Bulk ESS and Tail ESS were also sufficiently large for the reported parameters.

\end{document}